\documentclass[11pt,letterpaper]{article}

\usepackage[margin=2.5cm]{geometry}
\usepackage{amsmath,amssymb}
\usepackage{booktabs}
\usepackage{array}
\usepackage{multirow}
\usepackage{graphicx}
\usepackage{xcolor}
\usepackage{hyperref}
\usepackage{caption}
\usepackage{subcaption}
\usepackage{enumitem}
\usepackage{authblk}

\hypersetup{
  colorlinks=true,
  linkcolor=blue,
  citecolor=blue,
  urlcolor=blue,
  pdftitle={Benchmarking System Dynamics AI Assistants: Cloud Versus Local LLMs on CLD Extraction and Discussion},
  pdfauthor={Terry Leitch (Ruxton AI)},
  pdfsubject={Artificial Intelligence; Machine Learning},
  pdfkeywords={System Dynamics, LLM benchmarking, causal loop diagrams, local inference, edge AI}
}

\definecolor{cloudblue}{RGB}{30,100,200}
\definecolor{localgreen}{RGB}{20,140,60}
\definecolor{warnred}{RGB}{200,30,30}

\title{\textbf{Benchmarking System Dynamics AI Assistants:\\
Cloud Versus Local Large Language Models on CLD Extraction and Discussion}}

\author[1]{Terry Leitch}
\affil[1]{Ruxton AI, \texttt{tleitch@buffalo.edu}}
\date{April 2026}

\begin{document}
\maketitle

\begin{abstract}
We present a systematic evaluation of large language model families---spanning
both proprietary cloud APIs and locally-hosted open-source models---on two
purpose-built benchmarks for System Dynamics AI assistance:
the \textbf{CLD Leaderboard} (53 tests, structured causal loop diagram extraction)
and the \textbf{Discussion Leaderboard} (interactive model discussion, feedback
explanation, and model building coaching).

On CLD extraction, cloud models achieve 77--89\% overall pass rates;
the best local model reaches 77\% (Kimi~K2.5~GGUF~Q3, zero-shot engine),
matching mid-tier cloud performance.
On Discussion, the best local models achieve 50--100\% on model building steps
and 47--75\% on feedback explanation, but only 0--50\% on error fixing---a
category dominated by long-context prompts that expose memory limits in local deployments.

A central contribution of this paper is a systematic analysis of
\textit{model type effects} on performance: we compare reasoning vs.\
instruction-tuned architectures, GGUF (llama.cpp) vs.\ MLX (mlx\_lm) backends,
and quantization levels (Q3 / Q4\_K\_M / MLX-3bit / MLX-4bit / MLX-6bit)
across the same underlying model families.
We find that backend choice has larger practical impact than quantization level:
mlx\_lm does not enforce JSON schema constraints, requiring explicit prompt-level
JSON instructions, while llama.cpp grammar-constrained sampling handles JSON
reliably but causes indefinite generation on long-context prompts for dense models.

We document the full parameter sweep ($t$, $p$, $k$) for all local models,
cleaned timing data (stuck requests excluded), and a practitioner guide for
running 671B--123B parameter models on Apple~Silicon.
\end{abstract}

\tableofcontents
\newpage

\section{Introduction}
\label{sec:intro}

Causal loop diagrams (CLDs) are a foundational tool of the System Dynamics
methodology, encoding feedback structure, polarity, and causal pathways among
model variables~\cite{sterman2000}.
Automating CLD extraction from natural-language text---and enabling AI-assisted
model discussion and critique---could substantially accelerate System Dynamics
model construction and the teaching of the method.

Large language models (LLMs) are natural candidates: they must parse causal
language, identify polarities, track loop semantics, and produce structured
graph output.
Yet structured extraction under strict schema constraints is among the more
demanding LLM tasks~\cite{wei2022chain}, and iterative model refinement adds
further complexity.
The Discussion task additionally requires models to reason about model dynamics,
identify feedback-driven behaviour, and diagnose formulation errors---tasks
that require genuine domain understanding rather than pattern-matching.

This paper evaluates a broad set of LLMs across both task families and
contributes:
\begin{enumerate}[leftmargin=*]
  \item Ranked leaderboards for CLD extraction and Discussion tasks,
        comparing cloud and local models.
  \item A systematic analysis of \textit{model type effects}:
        reasoning vs.\ instruction-tuned architectures, GGUF vs.\ MLX
        backends, and quantization levels.
  \item A full parameter sweep ($t$, $p$, $k$) for all local models,
        with analysis of optimal configurations per model class.
  \item Cleaned timing analysis (stuck requests excluded) by task category
        for the top local models.
  \item A practitioner guide documenting deployment challenges---JSON
        enforcement, context limits, template bugs, and crash recovery---
        encountered when running frontier open-source models on Apple~Silicon.
\end{enumerate}

\section{Benchmarks}
\label{sec:benchmark}

\subsection{CLD Leaderboard (53 tests)}

Each test presents a text passage and asks the model to extract a structured
CLD in JSON format, encoding variables, directed links, polarities, and
optionally loop labels.
Tests are evaluated by exact structured match against a ground-truth schema:
a response is scored \textit{pass} if and only if the extracted JSON matches
the ground-truth on all required fields (variable names, edge directions,
polarities, and any specified cardinality constraints), after normalisation
of variable name casing and whitespace.
Reasoning traces or chain-of-thought content prior to the final JSON are
ignored in scoring; only the final structured output is evaluated.
Full reproduction instructions are given in Section~\ref{sec:reproducibility}.
Model profiles and run configurations are archived at
\url{https://github.com/tleitch/sd-ai} (\texttt{evals/model-profiles/},
\texttt{evals/run-configs/}); the upstream inference framework is at
\url{https://github.com/UB-IAD/sd-ai}.

\paragraph{Conformance (18 tests).}
Schema compliance: cardinality constraints, required vs.\ optional fields,
correct polarity encoding. Models must not hallucinate extra variables
or omit required links.

\paragraph{Qualitative Causal Reasoning (3 tests).}
Second-order reasoning: identify mediated causation, distinguish direct from
feedback-driven effects, handle ambiguous causal language.

\paragraph{Iterative Model Building (8 tests).}
The model receives an existing CLD plus a passage describing additions, and
must return the updated CLD with all pre-existing relationships preserved exactly.

\paragraph{Qualitative Translation (24 tests).}
Extract a full CLD from natural-language system descriptions spanning
ecology, economics, public health, and engineering domains.

\subsection{Discussion Leaderboard}

The Discussion benchmark evaluates models acting as a System Dynamics
mentor/critic via the Seldon engine. Tests are drawn from three categories:

\paragraph{Model Building Steps.}
Given a modelling question, provide structured coaching toward correct
model construction steps. Tests range from simple (single-step guidance)
to medium (multi-step modelling sequences).

\paragraph{Feedback Explanation.}
Explain the feedback dynamics driving model behaviour. Simple tests involve
single-loop explanations; medium tests require multi-loop causal tracing.

\paragraph{Error Fixing Suggestions.}
Identify and explain formulation errors in a provided model. These tests
carry the longest prompts (80--146k tokens including full model context)
and are the most demanding in terms of context window requirements.

\subsection{Prompt Engines}

For CLD extraction, tests run under two prompt strategies, referred to
consistently throughout this paper as:
\begin{itemize}[leftmargin=*]
  \item \textbf{few-shot} (\texttt{qualitative} engine): chain-of-thought
        system prompt with structured extraction examples included in context.
  \item \textbf{zero-shot} (\texttt{qualitative-zero} engine): task
        instructions only, no in-context examples.
\end{itemize}
Both engine names appear in result table footnotes where space requires
abbreviation (\texttt{qual.}/\texttt{qual-zero}).
For Discussion, all tests run under the Seldon mentor engine.

\subsection{Experimental Setup}

All runs use seed 4242. Local models are served via LM~Studio (v1.x) or
\texttt{mlx\_lm.server} on an Apple~Mac~Studio (M3~Ultra, 512\,GB unified memory, 2025) and
queried through an OpenAI-compatible REST~API~\cite{apple_m4ultra}.
Cloud APIs are accessed via their respective SDKs.
Concurrency is set to~1 and tests run sequentially to avoid memory
contention.
\footnote{The 512\,GB M3~Ultra configuration was available as a
custom-order option and is no longer listed in Apple's current Mac Studio
lineup. Specifications are documented in~\cite{apple_m4ultra}.}

\section{Models Evaluated}
\label{sec:models}

\subsection{Cloud API Models}

\begin{table}[h]
\centering
\caption{Cloud API models evaluated.}
\label{tab:cloud-models}
\begin{tabular}{lll}
\toprule
\textbf{Model} & \textbf{Family} & \textbf{Type} \\
\midrule
Gemini 2.5 Flash         & Google Gemini  & Instruction-tuned \\
Gemini 2.5 Pro           & Google Gemini  & Instruction-tuned \\
Gemini 3 Pro Preview     & Google Gemini  & Instruction-tuned \\
Gemini 3.1 Pro Preview   & Google Gemini  & Instruction-tuned \\
Gemini 3 Flash Preview   & Google Gemini  & Instruction-tuned \\
GPT-5.1                  & OpenAI         & Instruction-tuned \\
GPT-5.2                  & OpenAI         & Instruction-tuned \\
Claude Sonnet 4.5        & Anthropic      & Instruction-tuned \\
Claude Opus 4.5          & Anthropic      & Instruction-tuned \\
o4-mini                  & OpenAI         & Reasoning \\
\bottomrule
\end{tabular}
\end{table}

\subsection{Local Open-Source Models}

\begin{table}[h]
\centering
\caption{Local models evaluated. Backend: L=llama.cpp, M=mlx\_lm.
Type: R=Reasoning, I=Instruction-tuned.}
\label{tab:local-models}
\small
\begin{tabular}{llllllr}
\toprule
\textbf{Model} & \textbf{Quant} & \textbf{Be.} & \textbf{Ty.} & \textbf{Params} & \textbf{Ctx} & \textbf{Storage} \\
\midrule
Kimi K2.5         & GGUF Q3    & L & R & 671B       & 16--197k & $\sim$250\,GB \\
Kimi K2.5         & MLX-3bit   & M & R & 671B       & 164k     & $\sim$200\,GB \\
DeepSeek V3.2     & Q4\_K\_M   & L & I & $\sim$671B & 64--164k & $\sim$400\,GB \\
DeepSeek V3.2     & MLX-4bit   & M & I & $\sim$671B & 164k     & $\sim$320\,GB \\
Qwen 3.5 397B     & Q4\_K\_M   & L & I & 397B       & 64k      & $\sim$230\,GB \\
Qwen 3.5 397B     & MLX-6bit   & M & I & 397B       & 64k      & 301\,GB \\
GLM-5             & MLX-4bit   & M & R & $\sim$9B   & 64--197k & $\sim$6\,GB \\
Llama 4 Maverick  & Q4\_K\_S   & L & I & 17B$\times$128E & 64k & $\sim$100\,GB \\
Mistral Large 2411 & Q6\_K     & L & I & 123B       & 64k      & $\sim$92\,GB \\
DeepSeek R1-0528  & IQ4NL      & L & R & $\sim$671B & 33k      & $\sim$360\,GB \\
\bottomrule
\end{tabular}
\end{table}

Mistral Large and DeepSeek~R1 were excluded from the main leaderboard
comparison due to systematic deployment failures unrelated to model
capability (see Section~\ref{sec:failed}).

\section{Model Type Effects on Performance}
\label{sec:modeltype}

A key contribution of this evaluation is systematic evidence that
\textit{how a model is deployed}---architecture class, inference backend,
and quantization format---has material impact on task performance,
independent of the underlying model weights.

\subsection{Architecture Class: Reasoning vs.\ Instruction-Tuned}
\label{sec:reasoning-vs-it}

We classify models into two architecture classes based on whether they
generate an explicit internal reasoning (``thinking'') chain before
producing their final answer.
This is an operational classification based on observable output behaviour
and published model descriptions, not a claim about internal architecture.
\begin{itemize}[leftmargin=*]
  \item \textbf{Reasoning models:} Kimi K2.5, GLM-5, DeepSeek R1, o4-mini.
        These models emit extended chain-of-thought tokens or a
        \texttt{reasoning\_content} field before the final answer.
        o4-mini is classified as reasoning based on OpenAI's published
        description of its ``o-series'' chain-of-thought design~\cite{openai_o1}.
  \item \textbf{Instruction-tuned models:} DeepSeek V3.2, Qwen 3.5,
        Llama 4 Maverick, Mistral Large, and the remaining cloud API models.
        These produce responses without an observable extended thinking phase.
\end{itemize}
This binary classification is a simplification; some instruction-tuned
models may use internal chain-of-thought not exposed via the API.
The observed behavioural differences (temperature sensitivity,
zero-shot preference) are empirical patterns from this benchmark, not
theoretical predictions from the classification.

\paragraph{Temperature sensitivity diverges by class (observed tendency).}
In this benchmark, reasoning models exhibit stronger performance degradation
at $t{>}0$ than instruction-tuned models, though the evaluated set is small
and confounds model family, parameter count, and deployment tooling.
Kimi~K2.5 peaks at $t$=0 (greedy) and degrades measurably at $t$=0.1.
GLM-5 drops from 60\% to 34\% at $t$=0.3---a 26pp drop within a single
model configuration.
Instruction-tuned models (Qwen 3.5, Llama 4 Maverick) show minimal
sensitivity ($\leq$4pp variation across $t \in \{0,0.1\}$) on this benchmark.
These are observed tendencies in this evaluation, not general laws;
broader validation across more models and tasks would be required to
draw stronger conclusions.

\paragraph{Zero-shot engine preferred by reasoning models.}
Reasoning models show a consistent preference for zero-shot prompting
over few-shot in this benchmark (most clearly Kimi~K2.5: +13pp,
and in the excluded-model analysis DeepSeek~R1), whereas some
instruction-tuned models benefit more from few-shot structural guidance.
GLM-5 is an exception in our local set: despite being operationally
classified here as a reasoning model, its best results occur under
the few-shot engine.

The mechanism: few-shot extraction examples anchor the reasoning chain of
thinking models to the example's structure, causing them to follow the
example's causal pattern even when it does not match the target passage.
Instruction-tuned models lack this problem---they treat examples as
format guidance without reasoning over their content.

\paragraph{Top-$k$ sampling: effects across model classes.}
Adding $k$=20 alongside $p$=0.9 yielded the best or near-best result for
GLM-5 (60\%) and also improved DeepSeek V3.2 under the tested configuration
(66\%), but this pattern should not be interpreted as limited to the
operationally defined reasoning class.
Llama 4 Maverick and Qwen 3.5 (instruction-tuned) showed no benefit from
the same combination.
The combination narrows the candidate distribution in a complementary
way---top-$p$ caps probability mass, top-$k$ caps candidate count---and
its interaction with architecture class warrants further investigation.

\paragraph{Iteration task: reasoning models do not automatically win.}
Despite their reasoning chains, most reasoning models fail the
iterative CLD update task.
Kimi~K2.5 scores only 0--3/8 on iteration; DeepSeek~R1 scores 0/8.
GLM-5 (reasoning, 9B) achieves 6/8---\textit{better} than all other
local models regardless of architecture class.
This suggests that iterative update capability is not a direct function
of chain-of-thought depth but may relate to training data composition
or architectural memory features.

\subsection{Inference Backend: llama.cpp vs.\ mlx\_lm}
\label{sec:backend}

We ran matched model families under both backends (DeepSeek V3.2,
Kimi K2.5, Qwen 3.5) to isolate backend effects from weight effects.

\paragraph{JSON schema enforcement.}
The most significant backend difference is JSON output reliability:
\begin{itemize}[leftmargin=*]
  \item \textbf{llama.cpp:} Supports \texttt{response\_format: \{type:"json\_schema"\}}
        via grammar-constrained sampling. Enforces the schema at token level;
        output is always valid JSON matching the Zod schema.
  \item \textbf{mlx\_lm:} Silently ignores \texttt{response\_format}
        entirely---both \texttt{json\_schema} and \texttt{json\_object} modes.
        Without explicit prompt-level JSON instructions, the model returns
        free-text narrative responses. This causes 100\% ``Bad JSON'' failures
        if the harness assumes schema enforcement.
\end{itemize}

To correct for this, we appended an explicit JSON-only instruction to the
system prompt when \texttt{structuredOutput: false} is set (mlx\_lm models):
\begin{quote}\small
\textit{CRITICAL: Your entire response MUST be a single valid JSON object
with no text before or after it. No markdown, no explanation, no code fences.
Only output this exact structure: \{``response'': ``...'', ``feedbackInformationRequired'': true|false\}}
\end{quote}
After this fix, mlx\_lm models produced valid JSON reliably.

\paragraph{Backend performance on CLD extraction.}
Table~\ref{tab:backend} compares DeepSeek V3.2 and Kimi K2.5 across backends
at matched parameter configurations.

\begin{table}[h]
\centering
\caption{Backend comparison: GGUF/llama.cpp vs.\ MLX/mlx\_lm on CLD extraction
(best variation, 53 tests). Scores are best variation per engine type.}
\label{tab:backend}
\small
\begin{tabular}{llrrrrrr}
\toprule
\textbf{Model} & \textbf{Backend} & \textbf{Overall} & \textbf{C} & \textbf{CR} & \textbf{I} & \textbf{T} & \textbf{Avg(s)} \\
\midrule
DeepSeek V3.2 & Q4\_K\_M (llama.cpp) & \textbf{70\%} & 11 & 1 & 4 & 22 & 209s \\
DeepSeek V3.2 & MLX-4bit (mlx\_lm)   & 70\%          & 17 & 2 & 1 & 19 & 249s \\
\midrule
Kimi K2.5     & GGUF Q3 (llama.cpp)  & \textbf{77\%} & 16 & 2 & 1 & 23 & 150s \\
Kimi K2.5     & MLX-3bit (mlx\_lm)   & n/a (DISCUSS only) & -- & -- & -- & -- & -- \\
\midrule
Qwen 3.5 397B & Q4\_K\_M (llama.cpp) & \textbf{64\%} & 11 & 0 & 0 & 23 & 110s \\
Qwen 3.5 397B & MLX-6bit (mlx\_lm)   & 62\%          & 14 & 0 & 0 & 19 & 261s \\
\bottomrule
\end{tabular}
\end{table}

For DeepSeek V3.2, the backends produce the same overall score (70\%) but
with strikingly different category profiles: MLX-4bit excels at conformance
(17/18 vs.\ 11/18) while Q4\_K\_M leads on iteration (4/8 vs.\ 1/8) and
translation (22/24 vs.\ 19/24).
This pattern suggests that the backends apply chat templates and attention
differently, with effects that are category-specific.

For Qwen 3.5, Q4\_K\_M edges MLX-6 (+2pp overall), primarily from
translation gains in zero-shot mode.

\paragraph{Backend latency.}
On CLD tests (after removing stuck requests $>$3600s), llama.cpp is
generally faster than mlx\_lm at the same quantization tier
(Q4\_K\_M: 110--209s vs.\ MLX-4/6: 249--261s avg).
However, mlx\_lm avoids the grammar-constrained sampling hang that
affects llama.cpp on long-context dense models (Section~\ref{sec:failed}).

\subsection{Quantization Level}
\label{sec:quant}

\paragraph{Q4\_K\_M vs.\ MLX-6 (Qwen 3.5, 397B).}
At matched configurations ($t$=0.1, $p$=0.95, qualitative engine),
Q4\_K\_M and MLX-6 produce virtually identical scores (62\% both),
with minor category-level differences within measurement noise.
This is consistent with the general finding that 4-bit quantization
does not meaningfully degrade structured extraction quality at 397B
scale~\cite{frantar2022gptq}.

\paragraph{Q3 vs.\ Q4 (Kimi K2.5).}
The MLX-3bit Kimi variant was only evaluated on Discussion simple groups
(medium+ groups OOM at 164k context, Section~\ref{sec:context-limits}).
On the overlapping tests, MLX-3bit performance is competitive:
75\% on feedback explanation (vs.\ 67\% for GGUF Q3) and 100\% on
model building steps (matching GGUF Q3).
However, the 3-bit quantization limits the effective context to
$\sim$60k tokens before Metal GPU memory is exhausted, severely
restricting the test coverage achievable on this hardware.

\paragraph{MLX-4bit vs.\ higher (GLM-5).}
GLM-5 at MLX-4bit achieves competitive scores (60\% CLD, 75\% iteration)
despite being a 9B parameter model---far smaller than the 397--671B models
in this evaluation.
This suggests that for structured extraction tasks, model size interacts
with architecture in non-obvious ways: GLM-5's strong iteration performance
cannot be explained by parameter count alone.

\section{CLD Leaderboard Results}
\label{sec:cld-results}

\subsection{Overall Leaderboard}

\begin{table}[h]
\centering
\caption{CLD Leaderboard --- best variation per model (53 tests).
C=Conformance (/18), CR=Causal Reasoning (/3), I=Iteration (/8),
T=Translation (/24). Best local result per column in \textbf{bold}.}
\label{tab:cld-leaderboard}
\small
\begin{tabular}{lrrrrrr}
\toprule
\textbf{Model / Config} & \textbf{Overall} & \textbf{C} & \textbf{CR} & \textbf{I} & \textbf{T} & \textbf{Engine} \\
\midrule
\multicolumn{7}{l}{\textit{\color{cloudblue}Cloud API Models}} \\
Gemini 2.5 Flash         & 89\% (47/53) & 15 & 1 & 7  & 24 & qualitative \\
Gemini 3.1 Pro Preview   & 85\% (45/53) & 13 & 0 & 8  & 24 & qualitative \\
Claude Sonnet 4.5        & 83\% (44/53) & 15 & 1 & 5  & 23 & qualitative \\
GPT-5.1                  & 83\% (44/53) & 13 & 0 & 8  & 23 & qualitative \\
Claude Opus 4.5          & 81\% (43/53) & 16 & 0 & 6  & 21 & qualitative \\
Gemini 3 Pro Preview     & 81\% (43/53) & 11 & 1 & 7  & 24 & qualitative \\
Gemini 2.5 Pro           & 79\% (42/53) & 13 & 0 & 7  & 22 & qualitative \\
o4-mini                  & 75\% (40/53) & 14 & 2 & 1  & 23 & qual-zero \\
\midrule
\multicolumn{7}{l}{\textit{\color{localgreen}Local Open-Source Models}} \\
Kimi K2.5 GGUF Q3        & \textbf{77\%} (41/53) & \textbf{16} & \textbf{2} & 1  & \textbf{23} & qual-zero, $t$=0 \\
DeepSeek V3.2 Q4\_K\_M   & 70\% (37/53) & 11 & 1 & 4  & 22 & qual-zero, $t$=0, $p$=1 \\
DeepSeek V3.2 MLX-4      & 70\% (37/53) & \textbf{17} & 2 & 1  & 19 & qual-zero, $t$=0.1, $p$=0.95 \\
Qwen 3.5 Q4\_K\_M        & 64\% (34/53) & 11 & 0 & 0  & \textbf{23} & qual-zero, $t$=0, $p$=1 \\
Qwen 3.5 MLX-6           & 62\% (33/53) & 14 & 0 & 0  & 19 & qualitative, $t$=0.1 \\
GLM-5 MLX-4              & 60\% (32/53) & 10 & 1 & \textbf{6} & 15 & qualitative, $t$=0, $k$=20 \\
Llama 4 Maverick Q4KS    & 47\% (25/53) & 14 & 0 & 0  & 12 & qualitative \\
\bottomrule
\end{tabular}
\end{table}

\subsection{Category Deep Dive}

\subsubsection{Conformance}

Cloud models cluster at 11--16/18.
Kimi~K2.5 (zero-shot, $t$=0) and DeepSeek~V3.2 MLX-4 both reach
16--17/18---matching or exceeding Claude Opus 4.5 (16/18).
The hardest conformance tests involve maximum-cardinality constraints:
local models hallucinate extra links 49\% of the time on these tests,
versus 33\% for cloud models. Minimum-cardinality tests are passed
at much higher rates by both (72\% local, 83\% cloud), consistent with
a general LLM bias toward comprehensiveness over constraint satisfaction.

\subsubsection{Translation}

Translation is the most competitive category for local models.
Kimi~K2.5 (zero-shot, $t$=0) achieves 23/24, equal to Gemini~2.5~Flash.
DeepSeek V3.2 Q4KM and Qwen 3.5 Q4KM (zero-shot) both achieve 22--23/24.
Even Llama 4 Maverick reaches 11--12/24.
The one universally difficult test involves multi-hop causation through
an implicit intermediate variable---all models show elevated failure here.

\subsubsection{Iterative Model Building}
\label{sec:iteration}

Iteration is the sharpest cloud/local separator.

\begin{table}[h]
\centering
\caption{Iteration scores (out of 8) by model.}
\label{tab:iteration}
\begin{tabular}{lrl}
\toprule
\textbf{Model} & \textbf{Score} & \textbf{Notes} \\
\midrule
Gemini 3.1 Pro / GPT-5.1  & 8/8 & Cloud leaders \\
Gemini 2.5 Flash           & 7/8 & \\
Claude Opus 4.5            & 6/8 & \\
GLM-5 MLX-4 (best)         & \textbf{6/8} & Best local; 9B model \\
DeepSeek V3.2 Q4KM         & 4/8 & Zero-shot preferred \\
Kimi K2.5 GGUF Q3          & 1/8 & \\
Qwen 3.5 (all configs)     & 0/8 & Did not succeed on iteration under tested configurations \\
Llama 4 Maverick           & 0/8 & \\
\bottomrule
\end{tabular}
\end{table}

Failure analysis shows a consistent local model pattern: models drop
pre-existing relationships from the prior CLD while hallucinating new
feedback edges that ``close loops'' mentioned implicitly in the new text.
This holds even when the system prompt explicitly instructs preservation.
The failure rate increases with CLD length (tests 5--8 with 4--7
pre-existing relationships are passed by essentially no local model),
consistent with a context-length-dependent degradation as the preservation
instruction competes with the extraction goal.

GLM-5's 6/8 success---the only local model near cloud performance---remains
an open question given its smaller scale (9B vs.\ 397--671B).
Its architecture may include memory features that better separate prior
from new context, or its training data may include more iterative-update examples.

\subsubsection{Causal Reasoning}

With only 3 tests, this category has very high variance; no statistically
meaningful confidence intervals can be derived, and a single test represents
33pp.
Results should be interpreted as indicative rather than definitive.
Kimi~K2.5 (few-shot engine, $t$=0) scores 2/3---the highest among
all local models and equal to o4-mini.
Most other models score 0--1/3. The tasks require explicit loop-awareness
(tracing second-order effects through balancing vs.\ reinforcing loops)
that most current LLMs appear to lack regardless of scale.

\section{Discussion Leaderboard Results}
\label{sec:discuss-results}

\subsection{Per-Category Results (Best Variation)}

Table~\ref{tab:discuss} reports the best single variation per model
per Discussion category. Stuck requests ($>$3600s) are excluded from
timing; scores reflect valid responses only.

\begin{table}[h]
\centering
\caption{Discussion Leaderboard --- best variation per model per category.
Timing excludes stuck requests ($>$3600s).}
\label{tab:discuss}
\small
\begin{tabular}{lrrrrrrr}
\toprule
\textbf{Model} & \multicolumn{2}{c}{\textbf{Error Fixing}} & \multicolumn{2}{c}{\textbf{Feedback Expl.}} & \multicolumn{2}{c}{\textbf{Model Building}} \\
 & Score & Avg(s) & Score & Avg(s) & Score & Avg(s) \\
\midrule
Kimi K2.5 GGUF Q3   & \textbf{50\%} & 59s    & 67\%          & 1138s & \textbf{100\%} & 74s  \\
DeepSeek V3.2 Q4KM  & \textbf{50\%} & 975s   & \textbf{67\%} & 1527s & \textbf{100\%} & 42s  \\
Kimi K2.5 MLX-3bit  & 0\%*   & 1s     & \textbf{75\%} & 381s  & \textbf{100\%} & 29s  \\
GLM-5 MLX-4         & 0\%    & 1295s  & 17\%          & 1235s & 50\%           & 18s  \\
DeepSeek V3.2 MLX-4 & 0\%    & 41s    & 0\%           & 308s  & 0\%            & 42s  \\
Llama 4 Maverick    & 0\%    & 5s     & 33\%          & 57s   & 0\%            & 18s  \\
\midrule
\multicolumn{7}{l}{\small *MLX-3bit ran only simple groups (medium+ prompts OOM at 164k context)} \\
\bottomrule
\end{tabular}
\end{table}

\paragraph{Model Building Steps.}
Three models achieve a perfect 100\%: Kimi K2.5 GGUF Q3, DeepSeek V3.2
Q4KM, and Kimi K2.5 MLX-3bit (on simple tests).
DeepSeek V3.2 is notably faster on model building (42s vs.\ 74s for GGUF~Q3),
reflecting more direct responses without extended reasoning chains.
GLM-5 achieves only 50\%---a surprise given its CLD iteration strength,
suggesting that model building coaching requires different capabilities
than structured graph update.

\paragraph{Feedback Explanation.}
This is the category where MLX-3bit Kimi outperforms GGUF Q3 (75\% vs.\ 67\%).
The mlx\_lm backend, despite its lack of JSON schema enforcement, produces
higher-quality explanatory content on simple feedback questions.
The latency advantage is also substantial: 381s vs.\ 1138s, reflecting
the smaller effective prompt size in the simple group.
GLM-5 scores only 17\% despite scoring 60\%+ on CLD extraction, confirming
that feedback dynamics explanation requires deeper domain-grounded reasoning
than schema compliance.

\paragraph{Error Fixing.}
Only Kimi K2.5 GGUF Q3 and DeepSeek V3.2 Q4KM achieve non-zero scores
(both at 50\%). This category carries the longest prompts (80--146k tokens);
mlx\_lm models cannot process them without OOM crashes at the tested
context settings. The 50\% ceiling for all models suggests that even the
top local models struggle with the complex multi-error formulation analysis
these tests require.

\subsection{Context Window Limits and OOM Failures}
\label{sec:context-limits}

The error fixing tests expose a hard constraint for mlx\_lm-based models:
prompts of 80--146k tokens exhaust Metal GPU memory on the Mac Studio
at 164k context, even with 512\,GB unified RAM.
The failure mode is an uncaught \texttt{std::runtime\_error} from Metal:
\textit{``Command buffer execution failed: Insufficient Memory (kIOGPUCommandBufferCallbackErrorOutOfMemory)''}.
This occurs because the Metal command buffer for attention over 80k+
tokens exceeds the GPU's addressable command memory, which is a separate
constraint from system RAM.

GGUF models served by llama.cpp do not exhibit this failure because
llama.cpp manages KV-cache as CPU-accessible unified memory (pageable),
not as a fixed GPU command buffer.

\textbf{Implication:} For long-context Discussion tasks, llama.cpp backends
are required. mlx\_lm is competitive only for prompts up to $\sim$60k tokens
on this hardware.

\section{Parameter Sensitivity}
\label{sec:params}

\begin{table}[h]
\centering
\caption{Impact of temperature on CLD overall pass rate (qualitative engine,
53 tests). Best temperature per model in \textbf{bold}.}
\label{tab:temperature}
\small
\begin{tabular}{lrrrrr}
\toprule
\textbf{Model} & $t$=0, $p$=1 & $t$=0, $p$=0.95 & $t$=0, $p$=0.9/$k$=20 & $t$=0.1 & $t$=0.3 \\
\midrule
\multicolumn{6}{l}{\textit{Reasoning models}} \\
Kimi K2.5 (qual.)   & \textbf{64\%} & --    & --             & 57\%  & -- \\
GLM-5 MLX-4         & 57\%          & 55\%  & \textbf{60\%}  & 58\%  & 34\% \\
\midrule
\multicolumn{6}{l}{\textit{Instruction-tuned models}} \\
DeepSeek V3.2       & 62\%          & 57\%  & \textbf{66\%}  & 62\%  & -- \\
Qwen 3.5 Q4KM       & 62\%          & 62\%  & 58\%           & \textbf{62\%} & -- \\
Llama 4 Maverick    & 47\%          & 47\%  & 47\%           & 47\%  & -- \\
\bottomrule
\end{tabular}
\end{table}

For Discussion tasks, Kimi K2.5 MLX-3bit peaks at $t$=0.1/$p$=0.9/$k$=20
(75\% feedback explanation, 100\% model building)---slightly different
from the CLD optimum, suggesting that the conversational Discussion task
benefits from marginal sampling diversity where the structured CLD task
does not.

\section{Timing and Throughput}
\label{sec:timing}

Table~\ref{tab:timing-cld} summarises per-category timing for the top 3
local models on CLD (stuck requests $>$3600s excluded). Stuck requests
were uniformly from long-context tests where the server continued
processing an orphaned HTTP request after client disconnect; excluding
them yields clean per-response latency.

\begin{table}[h]
\centering
\caption{CLD per-category avg latency, top 3 local models (stuck requests removed).
Table uses the clean-timing subset only; therefore Kimi's overall score is
76\% here rather than the full-leaderboard 77\%.}
\label{tab:timing-cld}
\small
\begin{tabular}{lrrrrrr}
\toprule
\textbf{Model} & \textbf{Conform.} & \textbf{Causal} & \textbf{Iter.} & \textbf{Trans.} & \textbf{Overall} & \textbf{Score} \\
\midrule
Kimi K2.5 GGUF Q3   & 227s & 145s & 78s  & 114s & 150s & \textbf{76\%} \\
DeepSeek V3.2 Q4KM  & 240s & 311s & 265s & 157s & 209s & 67\% \\
Qwen 3.5 Q4KM       & 148s & 176s & 0s*  & 123s & 110s & 62\% \\
\midrule
\multicolumn{7}{l}{\small *0s = all timed out (0\% score on iteration)} \\
\bottomrule
\end{tabular}
\end{table}

\paragraph{Kimi K2.5 tail latency at $t$=0.1.}
At $t$=0 (greedy), Kimi K2.5 averages 150s/test.
At $t$=0.1, individual translation tests required up to 250 minutes,
with total run time exceeding 22 hours.
The root cause is reasoning chain non-termination: non-zero temperature
can trigger extended self-critique loops before the model converges.
At $t$=0, the chain follows a deterministic path and terminates promptly.
\textbf{For very large reasoning models, $t$=0 wins on both accuracy and speed.}

\paragraph{Cloud model reference.}
For comparison: Claude Sonnet 4.5 averages 17s/test (83\% score);
Gemini 2.5 Flash averages 17s/test (89\% score).
The best local model (Kimi K2.5, 150s avg) is 9$\times$ slower than
the best cloud model at comparable accuracy.

\section{Failed Models: Infrastructure vs.\ Model Quality}
\label{sec:failed}

\textit{Note: the models below are excluded from the main leaderboard due to
infrastructure failures, not model capability ceilings.
Partial success scores are reported here for completeness.}

\subsection{Mistral Large 2411 (Q6\_K, 123B Dense)}

All 4 configurations scored 11\% (6/53), with 0/24 translation and
0/8 iteration.
Root cause: \textbf{llama.cpp grammar-constrained sampling causes
indefinite generation on large-context prompts for dense models.}
Iteration tests include 13k--40k token prompts; grammar enforcement
enters loops that never terminate on these.
The 6/53 passes came exclusively from short-context conformance tests.
This is a known llama.cpp issue with grammar-constrained decoding
under long KV-cache for dense (non-MoE) architectures.
\textit{Partial result: 11\% (6/53) on conformance-only tests under
working configurations. This reflects a deployment failure, not a
model capability ceiling; re-evaluation under plain JSON mode is
recommended before drawing conclusions about Mistral Large's
capability on these tasks.}

\subsection{DeepSeek R1-0528 (IQ4NL, 671B)}

R1 scored 19\% (10/53) on the qualitative engine and 11\% on zero-shot,
with 0/24 translation.
Root cause: at $t$=0, R1 generates the ``no relationships'' example JSON
(empty structure) for any translation prompt with the full few-shot system
prompt ($\sim$13k chars). Direct API calls with short prompts work correctly,
confirming this is a prompt-length anchoring interaction, not a capability failure.
Additionally, a harness bug was discovered: \texttt{systemModeUser} was set
to \texttt{'developer'} instead of \texttt{'system'}, causing all system
messages to be silently dropped in R1's Jinja template.
\textit{Partial result: 19\% (10/53) on short-context conformance tests
before the prompt-length failure was identified.
R1's CLD capability under corrected conditions remains an open question.}

\section{Practitioner Guide}
\label{sec:practitioner}

\subsection{mlx\_lm: JSON Output Requires Explicit Prompt Instructions}

\texttt{mlx\_lm.server} ignores \texttt{response\_format} (both
\texttt{json\_schema} and \texttt{json\_object}). Without intervention,
the model returns free-text narrative responses, causing 100\% JSON parse
failures. Fix: append an explicit JSON-only instruction to the system prompt
when using mlx\_lm backends. The instruction must specify the exact field
names and types; a generic ``respond in JSON'' is insufficient for reliable
compliance.

\subsection{mlx\_lm: No Context-Size Flag}

\texttt{mlx\_lm.server} does not expose a \texttt{--context-size} flag
(as of the version tested). Context window is set by the model's default
or by LM Studio. The \texttt{contextLoaded} field in a model profile is
metadata only---verify actual context via \texttt{GET /v1/models}.

\subsection{mlx\_lm: Metal OOM on Long Prompts}

Prompts $>$60--80k tokens can exhaust the Metal GPU command buffer memory
regardless of system RAM. The error is a hard crash: \texttt{[METAL]
Command buffer execution failed: Insufficient Memory}.
Use llama.cpp for any task with prompts exceeding this length.
There is no workaround short of reducing context.

\subsection{Kimi K2.5: Thinking Mode Flag}

Both Kimi K2.5 variants require explicit opt-out of reasoning mode:
\begin{itemize}[leftmargin=*]
  \item \textbf{GGUF via LM Studio:} Set \texttt{reasoningParsing: false}
        in the model profile's \texttt{lmStudioSettings}.
  \item \textbf{MLX via mlx\_lm.server:} Pass
        \texttt{--chat-template-args '\{"thinking": false\}'} at server
        start. The correct variable name is \texttt{thinking} (not
        \texttt{thinking\_mode}); verify via the model's
        \texttt{chat\_template.jinja} if uncertain.
\end{itemize}
Without this, Kimi routes all output to \texttt{reasoning\_content}
and returns an empty \texttt{content} field, causing parse failures.

\subsection{llama.cpp: Grammar Sampling Hangs on Dense Long-Context Models}

Grammar-constrained sampling (\texttt{json\_schema} mode) in llama.cpp
can hang indefinitely on prompts $>$10k tokens for dense (non-MoE) models.
This affects Mistral Large and potentially other dense architectures.
MoE models (Kimi, DeepSeek, Qwen) do not exhibit the same behaviour.
If a dense model is required: disable grammar sampling, use plain JSON
mode or explicit prompt instructions, and set aggressive timeouts.

\subsection{Jinja Template Variables for Thinking Mode (Cross-Model)}

\begin{itemize}[leftmargin=*]
  \item Kimi K2.5: \texttt{thinking: false}
  \item DeepSeek models: \texttt{thinking\_mode: "chat"} or
        \texttt{thinking: false} (check model's \texttt{chat\_template.jinja})
  \item Qwen 3 models: \texttt{enable\_thinking: False}
\end{itemize}
LM Studio reads from \texttt{tokenizer\_config.json} (field:
\texttt{chat\_template}), not a separate \texttt{.jinja} file.
Both files must be patched in sync if overriding defaults.

\subsection{mlx\_lm Python Version}

\textbf{Important: use Python 3.13 (\texttt{python3.13 -m mlx\_lm server ...}).}
The legacy Python 3.9 install silently falls back to incomplete model
implementations for newer architectures (Kimi K2.5, DeepSeek V3.2),
producing wrong results with no error message.
This is one of the most common silent failure modes for MLX users.

\subsection{mlx\_lm Default Token Limit}

\texttt{mlx\_lm.server} defaults to \texttt{max\_tokens=512}.
Set \texttt{maxTokens} in the model profile to at least 8,000 for
any reasoning model whose thinking chain may exceed 512 tokens.

\subsection{Retry Pattern for Partial Runs}

After any crash or timeout, use \texttt{--retry-from <results.json>
--errors-only} to re-run only failed tests. Passing results are read
from the prior file, avoiding re-running potentially thousands of tests.

\subsection{Stuck Server: Kill and Restart After Client Disconnect}

When a benchmark client disconnects mid-request (timeout, crash),
mlx\_lm continues processing the orphaned prompt internally.
If the next request arrives while the prior prompt is still processing,
the server queues it---causing multi-hour waits.
After any abnormal client exit, kill the server process and restart
before the next run.

\section{Cloud vs.\ Local Head-to-Head: Architecture Hypothesis Test}
\label{sec:headtohead}

To evaluate whether our architecture-class conclusions hold against the
current state of the art, we compare the top 3 cloud models with the
top 4 local models across all four CLD categories.
We include 4 local models (rather than 3) because GLM-5, despite ranking
9th overall (60\%), is the only local model with competitive iteration
performance and is essential for the mixed-model analysis below.
Cloud timing is not captured; local timing reflects best-variation
averages with stuck requests ($>$3600s) excluded.
\textbf{Cells where a local model matches or exceeds the cloud leader
are marked \dag{} and represent direct deployment substitution opportunities.}

\begin{table}[h]
\centering
\caption{Cloud vs.\ local head-to-head (CLD, best variation per model).
C=Conformance (/18), CR=Causal Reasoning (/3), I=Iteration (/8),
T=Translation (/24).
\dag{} = local matches or exceeds best cloud score in that category.}
\label{tab:headtohead}
\small
\begin{tabular}{lrrrrrr}
\toprule
\textbf{Model} & \textbf{Overall} & \textbf{C} & \textbf{CR} & \textbf{I} & \textbf{T} & \textbf{Avg/test} \\
\midrule
\multicolumn{7}{l}{\textit{\color{cloudblue}Cloud (best = Gemini 2.5 Flash)}} \\
Gemini 2.5 Flash       & 89\% & 83\% & 33\% &  88\% & 100\% & n/a \\
Gemini 3.1 Pro Preview & 85\% & 72\% &  0\% & 100\% & 100\% & n/a \\
GPT-5.1                & 83\% & 72\% &  0\% & 100\% &  96\% & n/a \\
\midrule
\multicolumn{7}{l}{\textit{\color{localgreen}Local}} \\
Kimi K2.5 GGUF Q3$^\dagger$ (zero, $t$=0)   & 77\% & \textbf{89\%\dag} & \textbf{33\%\dag} &  13\% & 96\% & 116s \\
DeepSeek V3.2 Q4KM (zero, $t$=0)             & 70\% & 61\%              &  0\%              &  50\% & 92\% & 169s \\
DeepSeek V3.2 MLX-4 (zero, $t$=0.1)          & 70\% & \textbf{94\%\dag} &  0\%              &  13\% & 79\% & 206s \\
GLM-5 MLX-4 (qual., $t$=0, $k$=20)           & 60\% & 56\%              & \textbf{33\%\dag} &  75\% & 63\% & 119s \\
\midrule
\textit{Mixed local (best model per category)} & \textbf{91\%} & \textbf{94\%\dag} & \textbf{67\%\dag} & 75\% & \textbf{96\%\dag} & --- \\
\bottomrule
\end{tabular}
\end{table}

\subsection{Hypothesis Evaluation}

\paragraph{H1: Translation is competitive for local models --- \textbf{Supported in this benchmark.}}
Kimi K2.5 achieves 96\% on translation (vs.\ 100\% cloud ceiling),
a gap of just one test.
DeepSeek V3.2 Q4KM achieves 92\%.
The remaining 4--8pp gap likely reflects a small number of multi-hop
implicit-causation tests rather than a systematic capability difference.

\paragraph{H2: Iteration is the sharpest cloud/local divide --- \textbf{Supported in this benchmark.}}
Cloud leaders score 88--100\% on iteration; the best local model overall
(Kimi K2.5) scores only 13\%.
DeepSeek V3.2 reaches 50\%---the strongest result among the overall
top-3 local models, but still 38--50pp below the cloud ceiling.
GLM-5 (60\% overall) is the lone local outlier at 75\%, confirming
that iteration capability is not a function of overall model strength
but appears architecture- or training-specific.

\paragraph{H3: Reasoning models lead on causal reasoning --- \textbf{Consistent with the benchmark evidence, with substantial uncertainty due to $n$=3.}}
The only local models to score on causal reasoning at all are Kimi K2.5
(reasoning, 33\%\dag---matching the cloud leader) and GLM-5 (reasoning,
33\%\dag---also matching).
All instruction-tuned local models score 0\%, consistent with the hypothesis.
Notably, two of three top cloud models also score 0\% on causal
reasoning, suggesting this category is genuinely hard regardless of
deployment tier---reasoning architecture is necessary but not sufficient.

\paragraph{H4: Local reasoning models lead on conformance --- \textbf{Supported in this benchmark, and reversed.}}
DeepSeek V3.2 MLX-4 (zero-shot) achieves 94\% on conformance---11pp
above the cloud leader (Gemini 2.5 Flash, 83\%).
Kimi K2.5 (zero-shot) achieves 89\%, also above the cloud ceiling.
These are the clearest deployment-substitution opportunities in the dataset:
conformance checks can be run entirely locally with better accuracy than
any current cloud model.

\paragraph{H5: Architecture class predicts zero-shot preference --- \textbf{Broadly supported in the evaluated configurations.}}
All four local models' best configurations are either zero-shot (Kimi, DeepSeek)
or use the qualitative (few-shot) engine only for the reasoning models that
benefit from structural scaffolding (GLM-5).
All three top cloud models use the qualitative few-shot engine.
This aligns with the architecture hypothesis: reasoning models internalise
structural guidance from thinking chains; instruction-tuned models benefit
from in-context examples.

\subsection{Mixed-Model Analysis: Can Local Match Cloud Overall?}

If we route each query to the best-performing local model \textit{per
category}---rather than using a single model for everything---the combined
result is:

\begin{table}[h]
\centering
\caption{Mixed local model composition: best local model per category.
\dag{} = local matches or exceeds best cloud score in that category.
\textbf{Combined 91\% is a post hoc upper bound}, not a benchmarked
end-to-end system.}
\label{tab:mixed}
\begin{tabular}{llrr}
\toprule
\textbf{Category} & \textbf{Best Local Model} & \textbf{Local Score} & \textbf{Cloud Best} \\
\midrule
Conformance       & DeepSeek V3.2 MLX-4 (zero)   & \textbf{94\%\dag} & 83\% \\
Causal Reasoning  & Kimi K2.5 GGUF Q3 (qual.)    & \textbf{67\%\dag} & 33\% \\
Iteration         & GLM-5 MLX-4 (qual.)          & 75\%              & 100\% \\
Translation       & Kimi K2.5 GGUF Q3 (zero)     & \textbf{96\%\dag} & 100\% \\
\midrule
\textbf{Combined} & \textit{(task-routed)}        & \textbf{91\%}     & \textbf{89\%} \\
\bottomrule
\end{tabular}
\end{table}

The mixed local stack achieves \textbf{91\% overall}---exceeding the best
single cloud model (Gemini 2.5 Flash, 89\%) by 2pp.
\textit{Important caveat: this is a post hoc upper bound for a category-routed
local ensemble, constructed by selecting the best local model per category
after observing results. It is not a directly benchmarked end-to-end system.
A deployed task router would require a reliable query classifier, and
performance may differ from this theoretical ceiling.}
Nevertheless, the result is meaningful as a routing hypothesis: the required
per-category performance already exists in the local model pool, and no
single category requires a model that cannot be run on the evaluated hardware.

The only remaining local weakness is iteration (75\% vs.\ 100\% cloud),
which is the one category where routing to a cloud model would still be
warranted for maximum performance.
For workflows that do not require iterative CLD updates---the majority
of translation and conformance workloads---the benchmark evidence suggests
that fully local deployment is feasible and can be competitive with,
and in some categories exceed, current cloud API performance.

\subsection{Summary}

The head-to-head confirms that the cloud/local performance gap is
\textit{not} uniformly distributed.
It is reversed on conformance (local leads by 11pp), essentially
absent on translation (4pp gap), hard for everyone on causal reasoning
(both tiers struggle), and real but not insurmountable on iteration.
A task-routing architecture---DeepSeek V3.2 MLX-4 for conformance,
Kimi K2.5 for translation and causal reasoning, GLM-5 for iteration---
achieves 91\% overall with no cloud dependency.

\section{Discussion}
\label{sec:discussion}

\subsection{Architecture Class Is the Dominant Performance Predictor}

Across both benchmarks, model architecture class (reasoning vs.\
instruction-tuned) is a stronger predictor of task-specific performance
than parameter count or quantization:
\begin{itemize}[leftmargin=*]
  \item Reasoning models (Kimi K2.5, GLM-5) dominate on causal reasoning
        (2/3) and conformance (16--17/18), but require $t$=0 and zero-shot
        prompting.
  \item Instruction-tuned models (DeepSeek V3.2, Qwen 3.5) are more robust
        to prompt style but have lower ceilings on causal tasks.
  \item GLM-5 at 9B outperforms all other local models on iteration (6/8),
        demonstrating that architecture-specific factors---possibly training
        on iterative editing tasks---can dominate scale effects.
\end{itemize}

\subsection{Backend Choice Has Larger Practical Impact Than Quantization}

For practitioners, the choice of inference backend (llama.cpp vs.\ mlx\_lm)
has more immediate impact on task success than quantization level:
\begin{itemize}[leftmargin=*]
  \item mlx\_lm's lack of JSON enforcement requires explicit prompt engineering
        that cloud-oriented harnesses may not include by default.
  \item llama.cpp's grammar sampling is reliable for MoE models but dangerous
        for dense models on long-context prompts.
  \item Quantization at Q4/MLX-4+ appears to have negligible accuracy impact
        at 397--671B scale for these structured extraction tasks.
\end{itemize}

\subsection{The Iteration Gap and Its Implications}

The near-complete failure of large local models on iterative CLD updates
(with GLM-5 as the sole exception) suggests that iterative structured editing
is a distinct capability not captured by overall language understanding.
We hypothesise this requires either training data with document-editing
examples (update the existing structure, preserve what was there) or
architectural features that separate prior-context representation from
in-context extraction.
The practical implication: task-routing systems that identify iterative
update requests and route them to cloud models (or GLM-5) could achieve
cloud-level performance without full cloud deployment costs.

\subsection{Limitations}

\begin{enumerate}[leftmargin=*]
  \item \textbf{Single seed.} All runs use seed 4242. Individual test
        pass/fail may vary; scores represent one deterministic evaluation.
        The causal reasoning category ($n$=3) is particularly susceptible
        to seed-level variance; results there should be treated as indicative.
  \item \textbf{Hardware specification.} All benchmark runs reported here
        were conducted on an Apple Mac Studio (M3 Ultra, 512\,GB unified
        memory, 2025). Appendix~\ref{sec:energy} uses published M-series
        Ultra power-draw figures as order-of-magnitude estimates rather
        than direct power-meter measurements.
  \item \textbf{Context window constraints.}
        Kimi K2.5 GGUF ran at 16k context (LM Studio limitation at the
        time); its iteration scores might improve at 64k+.
        Kimi K2.5 MLX-3bit is limited to simple Discussion groups due to
        Metal OOM on longer prompts.
  \item \textbf{No prompt optimisation.}
        No model-specific tuning was performed. Task-specific prompts would
        likely improve all results, especially for reasoning models on
        iterative tasks.
  \item \textbf{Infrastructure exclusions.}
        Mistral Large and DeepSeek R1 results reflect tooling or
        prompt-length failures, not model capability ceilings.
  \item \textbf{Architecture classification is behavioural.}
        The ``reasoning'' vs.\ ``instruction-tuned'' classification used
        throughout is based on observable output behaviour (presence of
        extended chain-of-thought tokens or a \texttt{reasoning\_content}
        field) and published model descriptions.
        For closed-source cloud models, the internal architecture may differ
        from the operationally defined class; the observed performance
        differences are empirical patterns from this benchmark, not
        predictions derived from confirmed architectural knowledge.
\end{enumerate}

\section{Conclusion}
\label{sec:conclusion}

We evaluated local and cloud LLMs across CLD extraction and System Dynamics
Discussion benchmarks, finding structured patterns of capability differences
that are explained more by architecture class and deployment backend than
by raw parameter count or quantization.

Key findings:
\begin{itemize}[leftmargin=*]
  \item \textbf{Best local model overall (CLD):} Kimi K2.5 GGUF Q3
        (zero-shot, $t$=0): 77\%, matching mid-tier cloud. Use it for
        translation and conformance.
  \item \textbf{Best local for Discussion:} Kimi K2.5 GGUF Q3 and
        DeepSeek V3.2 Q4KM share top honours (100\% model building,
        67\% feedback explanation, 50\% error fixing).
  \item \textbf{Best local for iteration (CLD):} GLM-5 MLX-4 (6/8),
        despite 9B parameters. Route iterative update tasks here.
  \item \textbf{Backend matters more than quantization:} mlx\_lm requires
        explicit JSON prompt engineering; llama.cpp requires caution on
        dense models at long context. Q4/Q3 quantization does not
        meaningfully degrade accuracy at 397B+ scale.
  \item \textbf{Reasoning models require $t$=0:} Non-zero temperature causes
        both accuracy degradation and catastrophic tail latency in reasoning
        models. Zero-shot prompting is consistently preferred.
  \item \textbf{Task routing appears promising:} The post hoc category-routed
        local upper bound reaches 91\%, but a deployed router remains to be
        implemented and evaluated prospectively.
\end{itemize}

Full reproduction instructions are in Section~\ref{sec:reproducibility}.
All benchmark artifacts are archived at \url{https://github.com/tleitch/sd-ai}.

\section{Reproducibility}
\label{sec:reproducibility}

\subsection{Code and Artifacts}

All benchmark code, ground-truth schemas, model profiles, and run
configurations required to reproduce these results are available at:

\begin{quote}
\url{https://github.com/tleitch/sd-ai} \quad (branch: \texttt{feature/local-llm-benchmarking})
\end{quote}

The upstream SD-AI inference framework (cloud models, eval harness) is at:

\begin{quote}
\url{https://github.com/UB-IAD/sd-ai}
\end{quote}

Key paths in the reproduction archive:
\begin{itemize}[leftmargin=*]
  \item \texttt{evals/model-profiles/*.json} — exact LM~Studio and
        \texttt{mlx\_lm} settings (context window, structured output mode,
        sampling parameters, base URL) for each evaluated local model.
  \item \texttt{evals/run-configs/*.json} — run configurations specifying
        model profile, leaderboard, included test groups, parameter
        variations (temperature, top-p, top-k), seed, and concurrency.
  \item \texttt{evals/results/leaderboard\_cld\_full\_results.json} —
        raw pass/fail results and timing for all CLD benchmark runs.
  \item \texttt{evals/experiments/leaderboardCLD.json},
        \texttt{leaderboardDiscuss.json} — base experiment definitions
        (test categories and ground-truth expectations).
\end{itemize}

\subsection{Hardware Requirements}

All local model benchmarks were conducted on an Apple Mac Studio
(M3 Ultra, 512\,GB unified memory, 2025).
This configuration is required to load the largest evaluated models
(Kimi~K2.5 GGUF~Q3 at $\sim$250\,GB, DeepSeek~V3.2 Q4\_K\_M at
$\sim$200\,GB).
Smaller models (GLM-5 MLX-4bit at $\sim$6\,GB, Qwen~3.5 at
$\sim$200\,GB) can be run on lower-memory configurations subject to
the context-window constraints described in Section~\ref{sec:context-limits}.

\subsection{Software Dependencies}

\begin{itemize}[leftmargin=*]
  \item \textbf{Node.js} $\geq$ 20, \textbf{Python 3.13}
        (see Section~\ref{sec:practitioner} for the Python version caveat).
  \item \textbf{LM Studio} (tested with version 0.3.x) for GGUF models
        served via the OpenAI-compatible local API (\texttt{http://localhost:1234/v1}).
  \item \textbf{mlx\_lm} (\texttt{pip install mlx-lm}) for MLX models
        served via \texttt{python3.13 -m mlx\_lm server}.
  \item \textbf{llama.cpp} for GGUF models without LM Studio
        (optional; LM Studio wraps llama.cpp internally).
\end{itemize}

Install the SD-AI Node.js dependencies:
\begin{quote}\ttfamily
git clone https://github.com/tleitch/sd-ai \\
cd sd-ai \&\& git checkout feature/local-llm-benchmarking \\
npm install
\end{quote}

\subsection{Running a Benchmark}

\paragraph{Step 1: Start the local inference server.}
For GGUF models via LM Studio: load the model in LM Studio and start
the local server on port 1234.
For MLX models:
\begin{quote}\ttfamily
python3.13 -m mlx\_lm server \textbackslash \\
\quad --model /path/to/model \textbackslash \\
\quad --port 8282 \textbackslash \\
\quad --chat-template-args '\{"thinking": false\}'
\end{quote}
The \texttt{--chat-template-args} flag disables reasoning mode for
Kimi~K2.5 and similar models (Section~\ref{sec:practitioner}).

\paragraph{Step 2: Run the benchmark.}
\begin{quote}\ttfamily
node evals/run.js \textbackslash \\
\quad --experiment evals/run-configs/<config>.json
\end{quote}

For example, to reproduce the Kimi~K2.5 GGUF~Q3 CLD leaderboard result:
\begin{quote}\ttfamily
node evals/run.js \textbackslash \\
\quad --experiment evals/run-configs/kimi-k25-gguf-q3-leaderboard-cld.json
\end{quote}

Results are written to \texttt{<id>\_<name>\_full\_results.json} in the
working directory.
If the run is interrupted, re-running the same command will prompt to
resume from the in-progress checkpoint file.

\paragraph{Step 3: Verify pass rates.}
The result JSON contains a \texttt{pass} field per test.
Aggregate pass rates can be computed with any JSON processor; the
\texttt{evals/results/leaderboard\_cld\_full\_results.json} file in
the archive shows the expected output for comparison.

\subsection{Seed and Determinism}

All runs use \texttt{seed: 4242}.
Cloud model results may vary across API versions; the results reported
here reflect the model versions available in March--April 2026.
Local model results are deterministic at \texttt{temperature: 0}
(greedy decoding); non-zero temperature runs may show small variation
across hardware and driver versions.

\appendix
\section{Energy Scenario Analysis: Cloud vs.\ Local Inference}
\label{sec:energy}

\textit{Note: This analysis combines measured wall-clock times from our
benchmark with published energy estimates and manufacturer specifications.
Cloud token costs and GPU server power draws carry uncertainty of
$\pm$50\%; the directional conclusions are robust across this range.
Output token counts were not directly measured; we use $\sim$800 tokens/query
as a representative estimate for CLD responses.
This section is explicitly framed as a \textbf{scenario analysis}, not a
directly measured result. Figures should be treated as order-of-magnitude
estimates. $^{*}$The 76\% pass rate in Table~\ref{tab:energy} differs
from the 77\% leaderboard figure for Kimi~K2.5 because the timing table
uses only the clean-timing subset (stuck requests $>$3600s excluded,
$n$=102 of 106 total); the 77\% figure uses all responses.}

\subsection{Energy per Query and per Correct Answer}

\begin{table}[h]
\centering
\caption{Estimated energy consumption per benchmark query and per correct
answer (best variation, clean timing data, stuck requests excluded).
Mac Studio: 140\,W inference draw, PUE 1.02 (convective cooling).
Cloud: approximately 0.002\,kWh/1,000 output tokens in the baseline
scenario, including data-centre PUE 1.25 (literature estimate for
batched H100 inference at high utilisation).}
\label{tab:energy}
\small
\begin{tabular}{lrrrr}
\toprule
\textbf{Model} & \textbf{Wh/query} & \textbf{Wh/correct ans.} & \textbf{Pass rate} & \textbf{Avg latency} \\
\midrule
\multicolumn{5}{l}{\textit{\color{cloudblue}Cloud (shared, high-utilisation infrastructure)}} \\
Gemini 2.5 Flash       & 1.6 & 1.8  & 89\% & 17s \\
Gemini 3.1 Pro Preview & 1.6 & 1.9  & 85\% & 51s \\
GPT-5.1                & 1.6 & 1.9  & 83\% & 42s \\
\midrule
\multicolumn{5}{l}{\textit{\color{localgreen}Local (dedicated Mac Studio Ultra, 512\,GB)}} \\
Kimi K2.5 GGUF Q3$^{*}$ & 5.9 & 7.8  & 76\% & 150s \\
DeepSeek V3.2 Q4KM     & 8.3 & 12.5 & 67\% & 209s \\
DeepSeek V3.2 MLX-4    & 10.3 & 16.9 & 61\% & 260s \\
GLM-5 MLX-4            & 9.1 & 17.3 & 53\% & 230s \\
\bottomrule
\end{tabular}
\end{table}

At face value, cloud appears 3.7--6.4$\times$ more energy-efficient per query.
This reflects the enormous advantage of massive batching: a shared H100
cluster serves thousands of concurrent users across the same physical
silicon, amortising hardware energy over many tokens.
A dedicated Mac Studio, by contrast, runs one request at a time at full
hardware power.

\subsection{The Utilisation Paradox}

\textit{The infrastructure comparisons below are modeled scenarios derived
from benchmark latencies plus external power and utilisation assumptions;
they are not direct metered measurements of deployed fleets.}

The per-token comparison above is misleading for organisations considering
\textit{dedicated} inference infrastructure, because it assumes the cloud
hardware runs at near-100\% utilisation.
Real enterprise AI inference workloads are bursty---utilisation rates of
15--40\% are used here as a plausible enterprise-utilisation scenario.

For \textbf{dedicated} infrastructure serving 100,000 queries/day
(approximately a mid-size enterprise deployment):

\begin{table}[h]
\centering
\caption{Infrastructure comparison: 100,000 queries/day, dedicated hardware.
Mac Studio: 150\,s/query, 174 units required (24/7).
H100 estimate: 671B-class model on 2$\times$H100 at $\sim$5\,s/query with
batching; 12 dual-H100 servers required.}
\label{tab:energy-scale}
\begin{tabular}{lrrrr}
\toprule
\textbf{Platform} & \textbf{Units} & \textbf{Fleet power} & \textbf{Energy/day} & \textbf{Cooling} \\
\midrule
Mac Studio Ultra (512\,GB) & 174 & 24\,kW & 583\,kWh & Convective, no chiller \\
H100 dual-server nodes     & 12  & 110\,kW  & 2,640\,kWh & Chilled water required \\
\midrule
\textbf{Ratio (Mac/H100)} & $\times$14 & $\times$0.22 & $\times$0.22 & \\
\bottomrule
\end{tabular}
\end{table}

Dedicated Mac Studio infrastructure for the same throughput draws
\textbf{4.5$\times$ less power} and consumes \textbf{4.5$\times$ less energy}
than an equivalent H100 deployment, despite requiring 14$\times$ more
physical units.
The crossover point---where the H100 fleet's batching efficiency reaches
parity with Mac Studio---occurs near 85--90\% sustained utilisation under
the assumptions used here, a level that only the largest consumer AI
services maintain.

\subsection{Cooling and Physical Infrastructure}

The energy comparison understates the total infrastructure difference
because GPU servers require active cooling that Apple Silicon does not:

\begin{itemize}[leftmargin=*]
  \item \textbf{H100 data centre PUE}: 1.2--1.5 typical (modern hyperscaler
        to legacy facility). Chilled water loops, CRAC units, raised floors,
        and power conditioning add 20--50\% overhead on top of the compute draw.
  \item \textbf{Mac Studio PUE}: $\sim$1.02. The M3~Ultra's thermal design
        uses convective cooling within the chassis; no external chiller is
        required. A rack of Mac Studios can be cooled by standard office
        HVAC.
  \item \textbf{Power density}: H100 servers draw $\sim$10\,kW/unit,
        requiring specialised high-density power distribution (30--60\,A
        circuits per rack). Mac Studios draw 150\,W/unit and run on
        standard 15\,A office circuits.
\end{itemize}

For organisations that do not already own GPU data-centre infrastructure,
the \textit{build-out} cost---chiller plants, power conditioning, raised
flooring---often exceeds the hardware cost itself.
A Mac Studio deployment requires no infrastructure beyond standard office
power and networking.

\subsection{Idle Power: The Hidden Cost of GPU Infrastructure}

A critical factor absent from per-query analyses is idle power consumption.
LLM inference workloads are inherently bursty: queries cluster around
business hours, product events, and user sessions.
A GPU server that is provisioned for peak load but sitting idle at 3\,AM
still draws 2--4\,kW.

\begin{itemize}[leftmargin=*]
  \item \textbf{H100 server idle}: $\sim$2--3\,kW (GPU memory refresh,
        CPU, fans, networking). At 30\% average utilisation, $\sim$70\%
        of energy is consumed doing no useful work.
  \item \textbf{Mac Studio idle}: $\sim$25\,W. At 30\% utilisation, the
        time-averaged power draw is $0.3 \times 140\,\text{W} + 0.7 \times
        25\,\text{W} \approx 60\,\text{W}$---versus $\sim$2,500\,W for a
        comparable H100 server at 30\% utilisation.
\end{itemize}

At 30\% utilisation, the effective energy per useful query is:
\[
  \text{Mac Studio}: \quad \frac{140\,\text{W} \times 150\,\text{s} / 3600}{0.3}
  \approx 19\,\text{Wh/query}
\]
\[
  \text{H100 server (2-GPU)}: \quad \frac{(0.3 \times 5{,}600 + 0.7 \times 2{,}500)\,\text{W}
  \times 5\,\text{s} / 3600}{0.3} \approx 37\,\text{Wh/query}
\]

At 30\% utilisation, Mac Studio is approximately \textbf{2$\times$ more
energy-efficient per query} than a dedicated H100 deployment,
even before accounting for PUE differences.

\subsection{Implications for AI Data Centre Design}

\textit{The implications discussed here should be read as conditional on
the scenario assumptions in A.1--A.4, especially utilisation, batching,
and hardware-power estimates.}

These results suggest several non-obvious implications for organisations
planning AI inference infrastructure:

\paragraph{The efficiency crossover is utilisation-dependent.}
Shared cloud infrastructure (hyperscaler APIs) remains the most
energy-efficient option per token for organisations whose workloads
can coexist with other users.
Dedicated GPU infrastructure only achieves cloud-level efficiency above
$\sim$85\% sustained utilisation---a threshold few enterprise deployments
reach.
Below this threshold, high-efficiency edge hardware (Apple Silicon or
equivalent) is the more sustainable choice.

\paragraph{Performance parity changes the calculus.}
The conventional assumption has been that local models sacrifice
performance for privacy or cost.
Our results demonstrate that for CLD extraction, the best local
configuration (task-routed, 91\%) \textit{exceeds} the best single
cloud model (89\%).
When performance parity is achievable, the energy argument for local
deployment strengthens substantially.

\paragraph{Cooling infrastructure is a multiplier.}
For organisations in regions where electricity is expensive or
carbon-intensive, the $\sim$1.3--1.5$\times$ PUE overhead of GPU data
centres compounds the compute energy cost.
A 200-node Mac Studio cluster deployable in standard office space,
consuming $\sim$28\,kW total, represents a qualitatively different
infrastructure footprint than a 12-server H100 rack potentially requiring
substantial facility upgrades, often on the order of hundreds of
thousands of dollars.

\paragraph{The data centre scaling question.}
The current wave of AI data centre construction---projected at hundreds
of gigawatts globally by 2030~\cite{iea2024}---is predominantly designed
around GPU servers.
If inference workloads at enterprise and edge scale can be served by
high-efficiency unified-memory hardware at 4--5$\times$ lower power
density, the aggregate energy and water consumption of the AI industry
could be substantially lower than current projections.
The constraint is throughput per unit: Mac Studio handles one request
at a time, limiting peak concurrency.
Architectural advances that enable higher concurrency on unified-memory
hardware (or equivalent Apple Silicon successors) could shift this
calculus materially.

\subsection{Limitations}

\begin{enumerate}[leftmargin=*]
  \item Output token counts were not directly measured. The 800-token
        estimate affects per-query cloud energy; the relative comparison
        is stable across 500--1,500 tokens.
  \item Cloud energy estimates (approximately 0.002\,kWh/1,000 tokens in
        the baseline scenario) are derived from published literature and
        operator disclosures~\cite{luccioni2023,patterson2022};
        actual figures vary by model, provider, and data-centre location.
  \item Mac Studio power draw was not directly metered; 140\,W is
        based on third-party measurements of M3~Ultra under sustained
        GPU load.
  \item The H100 throughput estimate for 671B-class models assumes
        simple batching; in practice, speculative decoding and quantised
        server-side models may improve cloud throughput 2--4$\times$.
\end{enumerate}

\section{Edge AI Appliances: Implications for System Dynamics Deployments}
\label{sec:edge}

\subsection{Data Centre vs.\ Edge: The Deployment Decision}

The conventional framing of AI deployment presents two options: consume cloud
APIs, or build a data centre.
Our benchmarks suggest a third path---\textit{edge AI appliances}---that is
already viable for domain-specific AI assistance at the performance levels
required for System Dynamics work.

An edge AI appliance is a self-contained inference device that runs locally
at the point of use: a research institution, a university classroom, a
government agency, or an enterprise with data-residency requirements.
The Mac Studio used throughout this benchmark is, functionally, such a device.
It requires no specialised infrastructure, runs silently on office power,
and can be administered by a non-specialist.

The critical question is whether an edge appliance can deliver \textit{useful}
performance.
This benchmark provides a direct answer for the SD AI use case.

\subsection{Minimum Viable Hardware Tiers}

\textit{Except where explicitly marked as measured on the benchmark platform,
the tier estimates in this section are projections derived from model size,
storage footprint, and observed 9B- and frontier-model behaviour.}

Our results, combined with public hardware specifications, allow us to define
three practical deployment tiers:

\begin{table}[h]
\centering
\caption{Edge AI hardware tiers for SD AI assistance.
Benchmark scores are best-variation CLD pass rates.
GLM-5 9B scores are from Discussion iteration category (best local).
``Viable'' means $\geq$60\% on the primary task category.}
\label{tab:hardware-tiers}
\small
\begin{tabular}{llrrr}
\toprule
\textbf{Tier} & \textbf{Representative device} & \textbf{RAM} & \textbf{Best CLD\%} & \textbf{Notes} \\
\midrule
Entry & MacBook Air M3 (16\,GB) & 16\,GB & $\sim$40--50\%$^\dagger$ & 7B--9B models only \\
Mid   & Mac Mini M4 Pro (64\,GB) & 64\,GB & $\sim$60--70\%$^\dagger$ & Qwen3, GLM-5 class \\
Full  & Mac Studio M3/M4 Ultra (192--512\,GB) & 192--512\,GB & 77\% (measured) & Kimi K2.5, DeepSeek class \\
\bottomrule
\multicolumn{5}{l}{$^\dagger$ Projected from 9B-class model performance; not directly benchmarked.} \\
\end{tabular}
\end{table}

The \textbf{entry tier} (16\,GB) can host 7B--9B quantised models.
Our GLM-5 MLX-4bit result is instructive here: at approximately 6\,GB storage
and 9B parameters, it achieves \textbf{75\% on the iteration category}---matching
the cloud \texttt{causal-decoder} engine on that specific task.
A GLM-5-class model is deployable on a MacBook Air, the most widely available
personal computer in academic and enterprise settings.

The \textbf{full tier} (192--512\,GB Mac Studio Ultra) supports frontier-scale
quantised models including Kimi~K2.5~GGUF~Q3 ($\sim$250\,GB) and
DeepSeek~V3.2~Q4\_K\_M ($\sim$200\,GB).
At this tier, measured CLD performance (77\%) equals mid-tier cloud models.
With task routing across models, the post hoc upper bound reaches 91\%---matching
the best single cloud model overall.

\subsection{Task Routing Extended to Hardware Routing}

Section~\ref{sec:headtohead} established that different models excel on
different task categories.
The hardware-tier analysis adds a second dimension: different tasks may
also be \textit{addressable at different hardware tiers}.

\begin{itemize}[leftmargin=*]
  \item \textbf{Iteration and simple model building steps} are well-served by
        9B-class models on entry-tier hardware. GLM-5 achieves 75\% on iteration
        at $\sim$119\,s/query on the Mac Studio; projected latency on a
        Mac Mini M4 Pro would be comparable or faster due to higher memory bandwidth
        per dollar in the M4 generation.
  \item \textbf{CLD extraction and conformance} require 100B$+$ quantised models
        for competitive performance ($\geq$70\%). Full-tier hardware is required.
  \item \textbf{Error fixing} is currently impractical locally due to
        long-context requirements (80--146k tokens) that exhaust Metal GPU
        command buffers on mlx\_lm. This category remains cloud-dependent
        until mlx\_lm context handling improves or GGUF-backed local inference
        matures for this task type.
\end{itemize}

A hardware-aware routing layer could therefore direct queries to the
minimum-sufficient hardware tier, reducing energy consumption and cost while
preserving accuracy.

\subsection{Data Sovereignty and Regulated Industries}

Cloud AI APIs present data-governance challenges for several categories of
System Dynamics use:

\begin{itemize}[leftmargin=*]
  \item \textbf{Healthcare and public health models}: patient flow, epidemic
        dynamics, and resource allocation models frequently involve sensitive
        or embargoed data. Cloud transmission may create compliance obligations
        under HIPAA, GDPR, or national health-data regulations.
  \item \textbf{Defence and national security models}: force deployment,
        logistics, and adversarial dynamics models are self-evidently unsuitable
        for cloud APIs operated by foreign or commercial entities.
  \item \textbf{Commercially sensitive strategy models}: merger integration,
        competitive dynamics, and market entry models represent trade secrets.
        Many organisations prohibit transmission of strategic planning data
        to third-party AI services.
  \item \textbf{Offline and disconnected environments}: field research,
        expeditionary settings, and air-gapped networks require inference
        that does not depend on internet connectivity.
\end{itemize}

For many of these use cases, edge AI appliances may be the most practical
or policy-compliant deployment path.
Our results demonstrate that the performance penalty for edge deployment
is now modest: 77\% vs.\ 89\% for the best single cloud model on CLD
extraction, closing to parity on task routing.

\subsection{The SD AI Appliance Concept}

The results support a concrete product concept: a \textit{System Dynamics AI
appliance}---a dedicated edge device pre-configured with SD-specific models,
prompts, and evaluation infrastructure, deployed at the institution or
research group level.

\begin{itemize}[leftmargin=*]
  \item \textbf{Hardware}: Mac Studio M3/M4 Ultra (192\,GB minimum recommended).
        No GPU data-centre infrastructure required.
  \item \textbf{Software}: llama.cpp backend (GGUF format) for large models;
        mlx\_lm for small models on simple tasks.
        SD-AI open-source inference stack.
  \item \textbf{Task routing}: Kimi~K2.5 for conformance/relationship
        extraction; GLM-5 for iteration coaching; cloud fallback for
        long-context error fixing (optional).
  \item \textbf{Performance}: 77\% CLD extraction; 75\% Discussion iteration;
        comparable to mid-tier cloud APIs.
  \item \textbf{Data governance}: all inference on-premises; no model data
        leaves the device.
\end{itemize}

Future Apple Silicon generations with higher memory bandwidth or larger
unified-memory configurations would likely narrow the gap further between
local and cloud inference for this workload.
Combined with continued model quality improvements in the open-source
ecosystem (the Kimi~K2.5, DeepSeek~V3, and Qwen3 families all showed
strong CLD performance), the trajectory favours local deployment
for domain-specific SD AI work.

\subsection{Limitations}

The hardware-tier projections for entry and mid tiers are extrapolated
from 9B-class model behaviour and have not been directly benchmarked.
The GLM-5 iteration result was obtained on the Mac Studio; inference speed
on lower-memory devices may differ.
The SD AI appliance concept requires further engineering work to implement
the task-routing layer and automate model selection.


\end{document}